# A Study of Local Binary Pattern Method for Facial Expression Detection

Ms.Drashti H. Bhatt[1], Mr.Kirit R. Rathod[2], Mr.Shardul J. Agravat[3]

[1](Computer Engineering Dept., P G Student, C U Shah College of Engineering and Technology, India)
[2](Computer Engineering Dept, Asst. Professor, C U Shah College of Engineering and Technology, India)
[3](Computer Engineering Dept, Asst. Professor, C U Shah College of Engineering and Technology, India)

***ABSTRACT:*** *Face detection is a basic task for expression recognition. The reliability of face detection & face recognition approach has a major role on the performance and usability of the entire system. There are several ways to undergo face detection & recognition. We can use Image Processing Operations, various classifiers, filters or virtual machines for the former. Various strategies are being available for Facial Expression Detection. The field of facial expression detection can have various applications along with its importance & can be interacted between human being & computer. Many few options are available to identify a face in an image in accurate & efficient manner. Local Binary Pattern (LBP) based texture algorithms have gained popularity in these years. LBP is an effective approach to have facial expression recognition & is a feature-based approach.*

***Keywords –****Local Binary Patterns (LBP), Face Recognition, Expression Detection.*

## I. INTRODUCTION

Face detection is the process of locating the face in any particular image. The two important keys for any face detection algorithm are: what are the features to be taken from the training images, and what is the learning classifier or virtual machine to be used in the detection process. [9]

Goal of Face Detection is, in a given arbitrary image, to check whether the face is detected in figure or not. If they are present then it searched for the location where it is and extent of each face. It tells whether there is any human face, if there is, where it is. (Or where they are). [1]

Expression Detection can be classified in two ways that are for static images & dynamic images. [2] To derive that facial expression is a crucial step is to note that the facial expression detected is effective. Facial expression recognition is challenging & interesting problem in recent years. Facial Expressions are the best way to communicate & express the feelings. The important thing is how fluently we detect or extract the facial expression from image.

In next section we will discuss various facial expression strategies & categorize various methods for it. Section III contains the working method for LBP. Conclusion is specified in last section.

## II. RELATED WORK

**1. Face Detection Strategies:**

1.1 Appearance based methods: It includes approaches such as feature analysis, Graph matching, appearance based, direct correlation method, Eigenfaces method & FisherFaces method. Used in identifying fingerprinting, iris reorganization, hand geometry, voice etc. [10]

1.2 Feature Invarient based methods: Invariant features, unresponsive to different positions, brightness, and viewpoints, are utilized in this approach to detect human faces. A statistical model is usually built up for describing the relations among face features and the presence of the detected faces. Such face features, for instance, are Facial Features, Texture, and Skin Color. The feature-based detection has large computation and operates slowly.

1.3 Knowledge based methods: This method is aimed at finding invariant features of a face within a complex environment, thereby localizing the position of the face. Relationships among the features helpfully determine whether a human face appears in an image or not. [3]

1.4 Template based methods: A template cohering with human face features is used to perform a pattern-matching operation based on the template. Here the pre-defined templates are considered for eyes, lips eye brows etc. Then they are matched with the detected face. At last with comparison we get to know what exact is the expression of the face. [4]

**2. Expression Detection Strategies:**

2.1 LBP: It is a feature extraction technique. Has its excellence in Classification, clustering & segmentation. LBP features can be used as small patterns, which are balanced or regular with reference to monotonic grey scale revolution. [5]

2.2 POC: POC, which is sometimes called phase correlation, is one of the image matching techniques and has been successfully applied to biometric authentication and computer vision problems. The height and position of the correlation peak indicates difference between images. [6]

2.3 Haar Classifier: Haar Like features consists of two or three jointed black and white rectangles. The integral image is defined as the summation of the pixel values of the original image. The value at any location ($x$, $y$) of the integral image is the sum of the image's pixels above and to the left of location ($x$, $y$). The integral image value is calculated by finding out the difference between the sums of pixel gray level values within the black and white rectangular regions compared with raw pixel values. [7]

2.4 AdaBoost: AdaBoost algorithm is used to weight the selected weak classifier. All the weak classifiers are ranked to several cascades with the use of optimization process. Within each stage, an ensemble of several weak classifiers is trained using the AdaBoost algorithm.

2.5 Gabor Wavelet: Here the images represented by Gabor wavelets are chosen for its biological matter and technical properties. The Gabor wavelets are of similar shape as the receptive fields of simple cells in the primary visual cortex. We use Gabor wavelets (and not any other wavelets) for image representation because it represents the image based on the way the human mind does. This makes modeling computer vision based on human vision a more efficient and effective process. [8]

### III. LBP METHODOLOGY

The original LBP operator points the pixels of an image with decimal numbers, which are called *LBPs* or *LBP codes* that encode the local structure around each pixel. It proceeds as illustrated in Fig. 1: Each pixel is compared with its eight neighbors in a 3 × 3 neighborhood by subtracting the center pixel value. In the result, negative values are encoded with 0, and the others with 1. For each given pixel, a binary number is obtained by merging all these binary values in a clockwise direction, which starts from the one of its top-left neighbor. The corresponding decimal value of the generated binary number is then used for labeling the given pixel. The derived binary numbers are referred to be the LBPs or LBP codes. One limitation of the basic LBP operator is that they are unable to capture dominant features with large-scale structures. To deal with the texture at different scales, the operator was later generalized to use neighborhoods of different sizes. A local neighborhood is defined as a set of sampling points evenly spaced on a circle, which is centered at within the pixels are interpolated using bilinear interpolation, thus allowing for any radius and any number of sampling points in the neighborhood.

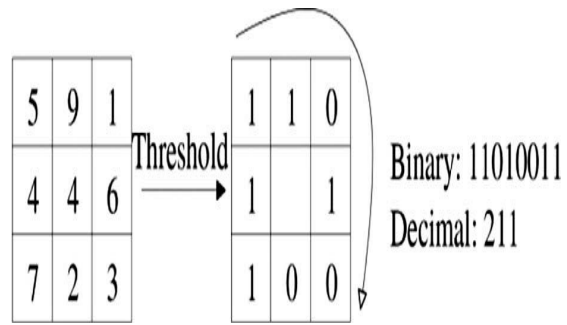

Fig: 1

Above details shows how the Local Binary Pattern (LBP) is working. It is the easy working & is converted in 3x3 matrixes. Local Binary Pattern (LBP) comparison can be shown through following table:

| | Accuracy | FPS | Detection Rate |
|---|---|---|---|
| Multi Class SVM | 84.1255% | 9.6 | Average |
| Corelations | 92% | 10 | Poor |
| Color & Feature bases | 92.3% | 19 | Good |
| Haar Classifiers | 94% | 12 | Good |
| LBP | 94.7% | 10 | Best |

Table: 1

## IV. CONCLUSION

Detection and Extraction of expressions from facial images is useful in many applications, such as robotics vision, video surveillance, digital cameras, security and human-computer interaction. The strategy that provides the best throughput with quite ease & best output is considered the most. Here LBP can be considered as the accurate pattern in finding the face & recognizing it. Accuracy factor is highest considering other factors. LBP is compatible with various classifiers, filters etc. We expect to develop an algorithm which provides us the best accuracy.